\documentclass[10pt,twocolumn,letterpaper]{article}

\usepackage[pagenumbers]{cvpr} % To force page numbers, e.g. for an arXiv version
\usepackage{makecell}
\usepackage{multirow}
\usepackage{graphicx}
\usepackage{xcolor}
\usepackage{booktabs}
\usepackage{array}
\usepackage[normalem]{ulem}
\usepackage[T1]{fontenc}
\usepackage{xcolor}
\usepackage{caption} % 提供 \captionsetup 和 \captionof 命令
\usepackage{afterpage}
\usepackage{float}
\definecolor{cvprblue}{rgb}{0.21,0.49,0.74}
\usepackage[pagebackref,breaklinks,colorlinks,allcolors=cvprblue]{hyperref}

\def\paperID{} % *** Enter the Paper ID here
\def\confName{CVPR}
\def\confYear{2026}

\title{$\text{F}^2\text{HDR}$: Two-Stage HDR Video Reconstruction via Flow Adapter \\
and Physical Motion Modeling}

\author{
Huanjing Yue$^{1,2}$ \quad Dawei Li$^{1,2}$ \quad Shaoxiong Tu$^{3}$ \quad Jingyu Yang$^{1}$\footnotemark[1]\\
$^1$School of Electrical and Information Engineering, Tianjin University \\
$^2$State Key Laboratory of Smart Power Distribution Equipment and System, Tianjin University\\
$^3$Huawei Technologies Co., Ltd.\\ 
{\tt\small \{huanjing.yue, dwli, yjy\}@tju.edu.cn, tushaoxiong1@hisilicon.com}
}

\newcommand{\best}[1]{\textbf{\textcolor{red}{#1}}}
\newcommand{\second}[1]{\textbf{\textcolor{blue}{#1}}}

\begin{document}
\maketitle
\begin{abstract}
Reconstructing High Dynamic Range (HDR) videos from sequences of alternating-exposure Low Dynamic Range (LDR) frames remains highly challenging, especially under dynamic scenes where cross-exposure inconsistencies and complex motion make inter-frame alignment difficult, leading to ghosting and detail loss. Existing methods often suffer from inaccurate alignment, suboptimal feature aggregation, and degraded reconstruction quality in motion-dominated regions. To address these challenges, we propose \textbf{$\text{F}^2\text{HDR}$}, a two-stage HDR video reconstruction framework that robustly perceives inter-frame motion and restores fine details in complex dynamic scenarios. The proposed framework integrates a flow adapter that adapts generic optical flow for robust cross-exposure alignment, a physical motion modeling to identify salient motion regions, and a motion-aware refinement network that aggregates complementary information while removing ghosting and noise. Extensive experiments demonstrate that $\text{F}^2\text{HDR}$ achieves state-of-the-art performance on real-world HDR video benchmarks, producing ghost-free and high-fidelity results under large motion and exposure variations. Our code is available at \url{https://github.com/wei1895/F2HDR}.
\end{abstract}

\renewcommand{\thefootnote}{\fnsymbol{footnote}}
\footnotetext[1]{This work was supported in part by the National Natural Science Foundation of China under Grant 62472308 and Grant 62231018. Corresponding author: Jingyu Yang}
\section{Introduction}
\label{sec:intro}

Compared to Low Dynamic Range (LDR) videos, High Dynamic Range (HDR) videos offer richer details and superior visual quality. However, constrained by the hardware capabilities of consumer-grade cameras, the dynamic range of real-world scenes often far exceeds the recording range of image sensors, resulting in detail loss in both overexposed and underexposed regions. Unlike HDR video acquisition methods that rely on specialized hardware~\cite{nayar2000high, mcguire2007optical, huggett2009dual, rebecq2019high, han2023hybrid}, reconstructing HDR frames from sequences of LDR frames with alternating long and short exposures is a more cost-effective approach. Furthermore, HDR content also facilitates downstream high-level vision tasks such as object detection~\cite{onzon2021neural} and segmentation~\cite{martinez2017image}.

\begin{figure}[t]
\centering
{\includegraphics[width=0.9\linewidth,clip, trim=1mm 1mm 1mm 1mm]
{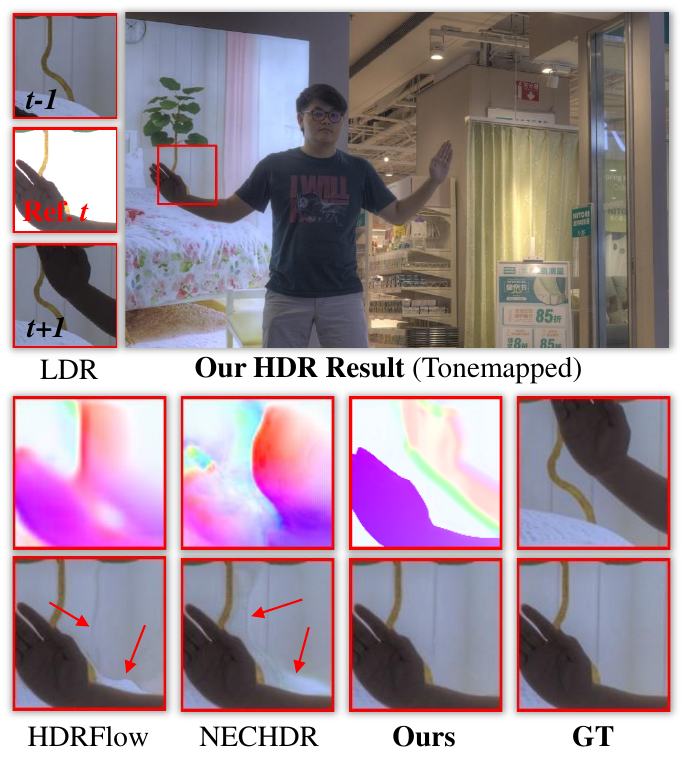}}
\caption{Comparison on the real-world HDR video reconstruction benchmark~\cite{shu2024towards} with state-of-the-art methods~\cite{xu2024hdrflow, cui2024exposure}. The challenging scene involves both global and local motion, while the reference frames suffer from severe information loss. Our \textbf{$\text{F}^2\text{HDR}$} achieves the most robust optical flow estimation and the most accurate detail restoration.}
\label{fig:teaser}
\vspace{-15pt}
\end{figure}

Multi-frame based HDR reconstruction methods can be generally categorized into two types: single-stage and two-stage approaches. The single-stage methods \cite{kalantari19, xu2024hdrflow,yue2025staggered} follow the alignment–fusion paradigm, directly merging the aligned LDR frames to obtain the final HDR result. Such networks are characterized by high computational efficiency and fewer parameters. However, achieving precise alignment across frames with varying exposures is challenging for dynamic scenes. LDR inputs often exhibit incomplete information since long-exposure frames are prone to saturation while short-exposure frames fail to preserve details in dark regions. Consequently, obtaining accurately aligned frames becomes difficult, especially in the presence of occlusions or fast-moving objects. These limitations frequently lead to ghosting and color-shift artifacts in the fused HDR outputs, particularly in saturated and low-light areas. To address this issue, two-stage networks~\cite{chen2021hdr,shu2024towards, cui2024exposure,  chen2025ultrafusion, yan2024dynamic,yue2023hdr} are proposed, which divides the reconstruction process into coarse HDR reconstruction (or coarse aligned result) and HDR refinement stages. The coarse stage focuses on recovering global structures and luminance, providing a stable foundation for subsequent refinement. The refinement stage alleviates the misalignment of the previous stage and achieves better feature aggregation to mitigate ghosting artifacts. Compared to single-stage methods, two-stage approaches typically yield better HDR reconstruction by leveraging the complementary strengths of the coarse and refinement stages. However, there are still two challenges to be solved.   

\textit{How to generate well-aligned frames in the first stage?} Current two-stage methods usually adopt simple alignment strategy in the first stage, such as optical flow based alignment \cite{kalantari19, chen2021hdr, chen2025ultrafusion,cui2024exposure} or global motion estimation \cite{shu2024towards}, which cannot generate well aligned results and the errors will propagate to the second stage. We observe that the optical flow models pretrained on extensive optical flow datasets often struggle to align occluded regions across varying exposures in HDR scenarios \cite{chen2021hdr, chen2025ultrafusion}. On the other hand, the retrained task-oriented flow models ~\cite{xu2024hdrflow, cui2024exposure} cannot generate fine edge details. The main reason is that there is no suitable flow dataset for HDR reconstruction and the current flow network is trained by HDR reconstruction loss rather than flow prediction loss. Although the pretrained flow network can generate offset with fine edges but they cannot deal with over-exposed or under-exposed regions well. To leverage the advantages of pretrained optical flow and meet cross-exposure alignment requirements, we propose to learn residual flow via a learnable flow adapter. We first utilize a pretrained optical flow model to generate a coarse flow with well-preserved edge details, and then refine it with the residual flow to make it adapt to HDR reconstruction. As illustrated in Fig.~\ref{fig:teaser}, our flow adapter generates accurate flow offsets with clear edges for the over-exposed regions.  

\textit{How to refine the artifacts in the second stage?} The first stage can generate a coarse HDR result, which still may suffer from ghosting and noise in the motion regions. Therefore, the main task in the second stage is to refine these artifacts. Previous methods usually fuse the same modality images to generate the final result. For example, Chen \textit{et al.} fed the coarse HDR results into a fine-scale alignment and fusion network to generate the final HDR result \cite{chen2021hdr}. Yan \textit{et al.} fed the corrected LDRs into the second stage for HDR generation \cite{yan2024dynamic}.  The work in \cite{chen2025ultrafusion} fed the aligned LDR result into the second stage for inpainting and reconstruction. Different from them, the work in \cite{cui2024exposure} fed the original LDRs, generated LDRs, and coarse HDR features into the second-stage blending network. By using the three modalities, it achieves better HDR reconstruction results. However, there is no guidance for the blending process, which cannot handle the difficult regions in a more targeted manner. Shu \textit{et al.} \cite{shu2024towards} 
introduced an exposure mask into the second stage. However, the mask mainly depicts the bright regions rather than difficult regions. Therefore, in this work, we propose physical motion modeling to guide the fusion process. Considering that our flow obtained in the first stage depicts the motion regions, we extract the motion mask based on the flow, and then utilize the motion mask to modulate the features extracted from aligned LDR features. These features are further fused together with the coarse HDR result to generate the final HDR result. Since our motion mask is extracted from the flow, the optimization of the second stage can also optimize the flow result, which further leads to a better alignment result in the first stage. 
In summary, our contributions are as follows. 

\begin{itemize}
\item We propose a two-stage HDR video reconstruction framework named $\text{F}^2\text{HDR}$, which utilizes the flow information in two different manners. In the first stage, we propose a flow adapter based on pretrained flow models, which generate robust flow for alternating exposures. 
\item In the second stage, we propose a physical motion modeling module based on the flow to predict the regions with noise and ghosting artifacts. The aligned LDR features are first modulated by the mask, and then fused with the coarse HDR result after enhancement. The second stage selectively aggregates reliable cues from different sources to compensate for missing or corrupted details. 
\item Extensive quantitative and qualitative experiments demonstrate that $\text{F}^2\text{HDR}$ achieves \textbf{state-of-the-art} performance on real-world HDR video benchmarks.
\end{itemize}

\section{Related Work}
\label{sec:formatting}
\textbf{Multi-Exposure HDR Image Reconstruction.}
The most popular paradigm in HDR image reconstruction is to merge multiple images captured under different exposure settings. Most single-stage HDR reconstruction approaches perform a single Alignment-Fusion operation~\cite{kalantari2017deep, catley2022flexhdr, kong2024safnet, liu2021adnet, yan2019attention} or rely solely on feature-level fusion~\cite{tel2023alignment, yan2020deep, ye2021progressive} to produce an HDR-domain output. These methods typically design convolutional or attention-based architectures to fuse multi-exposure information. However, due to inevitable information loss and misalignment during the registration process, such approaches often suffer from noticeable artifacts.
To achieve more precise alignment, two-stage HDR reconstruction frameworks~\cite{kong2024safnet,li2025afunet} adopt a coarse-to-fine design that progressively refines the HDR results, demonstrating stronger reconstruction performance.
Recent foundation model-based works~\cite{chen2025ultrafusion, yan2024dynamic} formulate multi-exposure HDR reconstruction as a two-stage segmentation–inpainting pipeline. Yan et al.~\cite{yan2024dynamic} use SAM and Stable Diffusion to handle motion and misalignment, but such methods demand heavy computation and suffer from low fidelity and poor temporal consistency.

\noindent 
\textbf{HDR Video Reconstruction.}
Hardware-based HDR video methods~\cite{tocci2011versatile,kronander2014unified,mcguire2007optical,heide2014flexisp,choi2017reconstructing} use specialized designs but are too complex for widespread deployment. A more accessible line of work reconstructs HDR videos from LDR sequences captured with alternating exposures. Traditional methods are built upon explicit motion compensation, where inter-frame motion is modeled and corrected using hand-crafted alignment schemes~\cite{kang2003, mangiat2010high, kalantari13}. With the emergence of deep learning, Kalantari \etal~\cite{kalantari19} proposed a CNN framework that employs optical flow for alignment and learns pixel-wise fusion weights for HDR synthesis.

Recent studies have explored single-stage HDR video reconstruction from various perspectives. Chung et al.~\cite{lan_hdr} employ a luminance-based attention mechanism for alignment and a dedicated hallucination module to handle saturated regions. Xu et al.~\cite{xu2024hdrflow} design a task-specific optical flow network, achieving lightweight and real-time HDR video reconstruction. Guan et al.~\cite{guan2024diffusion} introduce a latent diffusion framework that incorporates exposure-aware conditioning and temporal alignment for HDR reconstruction. However, such single-stage designs still struggle to achieve robust alignment under complex motion and exposure variations, and accurate optical flow estimation across differently exposed frames remains a challenging problem.

Chen et al.~\cite{chen2021hdr} were the first to propose a coarse-to-fine two-stage framework for HDR video reconstruction, which performs coarse fusion within grouped frames followed by feature-domain alignment and refinement using deformable convolutions. In a different approach, Cui et al.~\cite{cui2024exposure} first reconstruct exposure-compensated reference frames and coarse HDR results through a feature pyramid network, and then conduct a second-stage weighted fusion to enhance temporal consistency.
However, existing two-stage HDR video reconstruction methods remain limited by alignment accuracy in the first stage, and often overlook the role of motion cues in guiding HDR detail refinement during the second stage, which may lead to error accumulation and inefficient computation.

Unlike previous works, our method introduces a lightweight Flow Adapter that effectively transforms the predictions of a general optical flow model into task-adaptive flow fields tailored for HDR video reconstruction.
This design enables our framework to obtain more accurate HDR results and precise motion cues already in the first stage. Furthermore, we introduce motion-guided feature modulation to direct the network’s attention toward regions with a high likelihood of containing artifacts.

\section{Method}

\begin{figure*}[t]

    \centering
    \includegraphics[width=0.85\linewidth]{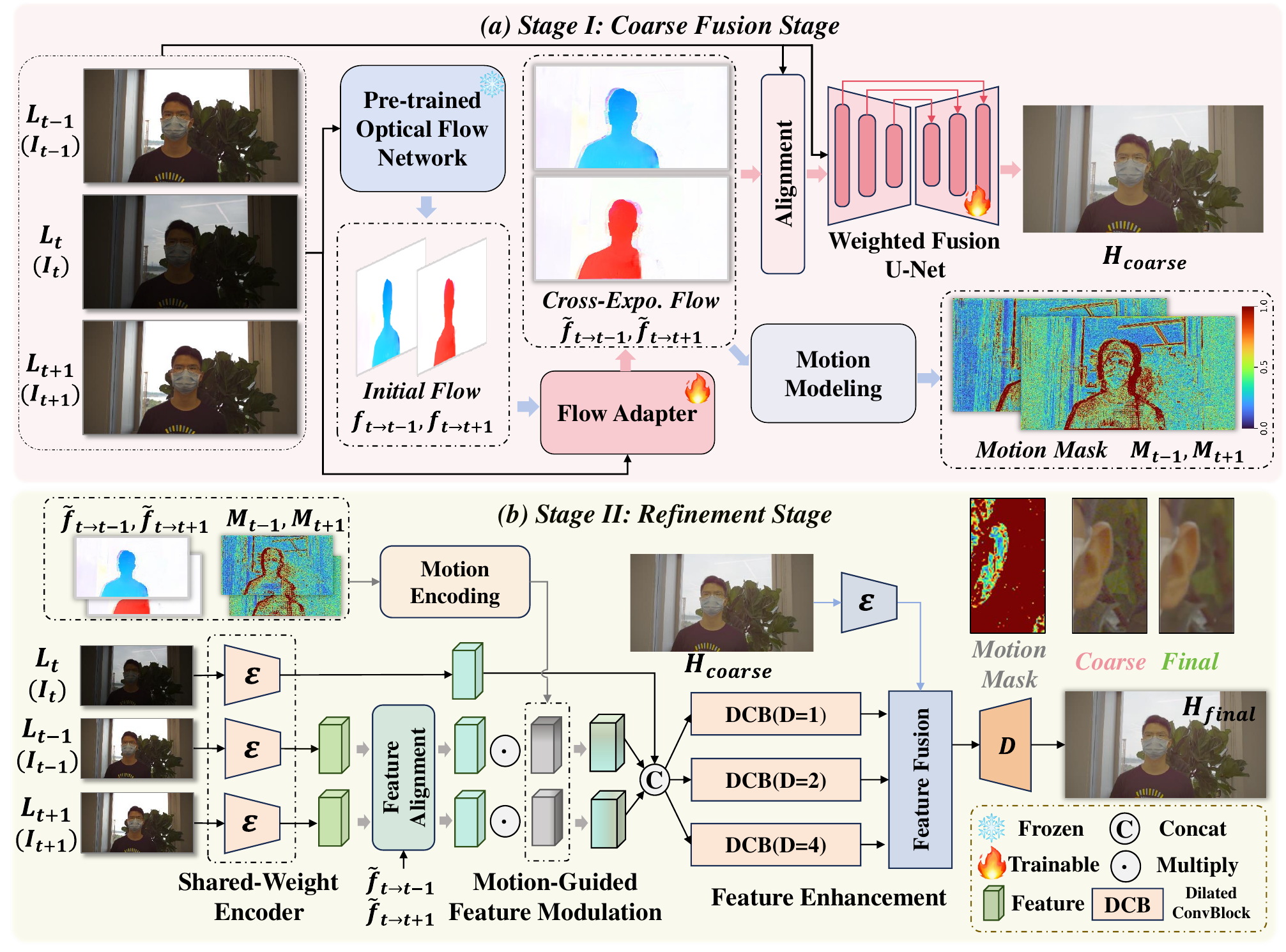}
    \caption{Network architecture of the proposed $\text{F}^2\text{HDR}$, which includes coarse fusion stage and refinement stage. 
    }
    \label{fig:network}
    \vspace{-15pt}
\end{figure*}

Capturing video frames with two alternating exposures is more feasible than that with multiple alternating exposures. Therefore, in this work, we take the LDR video sequence $\{L_t|t=1,2,...,n\}$ with two alternating exposures as input. We aim to reconstruct high quality HDR frames $\{H_t|t=1,2,...,n\}$ 
by fusing neighboring LDR frames. The middle frame $L_t$ is selected as the reference, and its two neighbors $L_{t-1}$ and $L_{t+1}$  are used to assist the reconstruction process.  

The overall framework of the proposed $\text{F}^2\text{HDR}$ is illustrated in \cref{fig:network}. It consists of two stages. The first stage is coarse fusion stage, which includes flow estimation, weighted fusion, and motion modeling. The second stage is refinement stage, which includes feature modulation, enhancement, and fusion modules. At the coarse fusion stage, we first estimate the optical flows $f_{t\rightarrow t-1}$ and  $f_{t\rightarrow t+1}$ via our proposed flow adapter. Then, we align the HDR domain input $I_{t-1}$ and $I_{t+1}$ obtained from $L_{t-1}$ and $L_{t+1}$ with the reference frame $I_{t}$ based on the flow, and the aligned frames are fused together to generate  a coarse HDR result $H_{\text{coarse}}$. Meanwhile, we extract the physical motion information from the flow via a motion modeling module. At the refinement stage, we extract features from the LDR input, which are aligned with the reference based on the flow $f_{t\rightarrow t-1}$ and  $f_{t\rightarrow t+1}$. Hereafter, these aligned features are modulated by the coded motion features. Then these features are enhanced and fused with the feature extracted from the coarse HDR result, after going through a decoder leading to the final HDR result $H_{\text{final}}$. 

Following previous works, we map each LDR frame to the linear HDR domain through gamma correction:
\begin{align}
\label{eq:ldr_to_hdr}
I_t = \frac{L_t^{\gamma}}{e_t},
\end{align}
where $e_t$ is the exposure for frame $L_t$ and $\gamma=2.2$.
\subsection{Stage I: Coarse Fusion Stage}
\subsubsection{Flow Estimation with Flow Adapter}
Considering that the pretrained optical flow network cannot adapt to HDR scenarios, and the retrained flow network cannot produce flows with clear edges. As shown in Fig.~\ref{fig:teaser}, the flow maps generated by the retrained flow network in \cite{cui2024exposure,xu2024hdrflow} are blurry. In contrast, we propose a flow adapter to predict residual flow, which refines the pre-trained flow prior and tailors it for HDR scenarios.

Given three consecutive LDR frames $\{L_{t-1}, L_t, L_{t+1}\}$ with their respective exposures $\{e_{t-1}, e_t, e_{t+1}\}$, we first normalize the exposure of the reference frame $L_t$ to match its neighboring frames. In this way, we can get two inputs with the same exposure, which is the setting of pretrained optical flow networks. Specifically, 
\begin{align}
g_{t\rightarrow t-1}(L_t) &= 
\text{clip}\!\left(\!\left({(L_t^{\gamma}}/{e_t}) e_{t-1}\right)^{\!\!1/\gamma}\!\right), \nonumber\\
g_{t\rightarrow t+1}(L_t) &= 
\text{clip}\!\left(\!\left({(L_t^{\gamma}}/{e_t}) e_{t+1}\right)^{\!\!1/\gamma}\!\right),
\end{align}
where $\gamma=2.2$ is the gamma correction coefficient.
Then, $\{L_{t-1}, g_{t\rightarrow t-1}(L_t)\}$ is fed into the pretrained optical flow network SEA-RAFT \cite{wang2024sea} to obtain the initial flow $f_{t\rightarrow t-1}$. Similarly, $f_{t\rightarrow t+1}$ is obtained for $\{L_{t+1}, g_{t\rightarrow t+1}(L_t)\}$. The initial flow is not good for over-exposed or under-exposed regions. Therefore, we further refine it with a flow adapter.  
The adapter is constructed by a shallow residual CNN that refines the coarse flows using both image content and motion cues. 
Given three consecutive LDR frames $\{L_{t-1}, L_t, L_{t+1}\}$ and the coarse flows $f_{t\rightarrow t-1}$ and $f_{t\rightarrow t+1}$, we first form the concatenated input
\begin{align}
\mathbf{x}_t = \big[L_{t-1},\, L_t,\, L_{t+1},\, f_{t\rightarrow t-1}/\lambda,\, f_{t\rightarrow t+1}/\lambda\big],
\end{align}
where $\lambda = 20$ is a fixed normalization factor.  
The adapter consists of several Conv–PReLU blocks with different dilation rates and predicts residual flows
\begin{align}
\big[\Delta f_{t\rightarrow t-1},\, \Delta f_{t\rightarrow t+1}\big] 
= \mathcal{A}(\mathbf{x}_t),
\end{align}
where $\mathcal{A}(\cdot)$ denotes the flow adapter.  
The refined flows are obtained by residual correction:
\begin{align}
\tilde{f}_{t\rightarrow t-1} &= f_{t\rightarrow t-1} + \lambda\,\Delta f_{t\rightarrow t-1}, \\
\tilde{f}_{t\rightarrow t+1} &= f_{t\rightarrow t+1} + \lambda\,\Delta f_{t\rightarrow t+1}.
\end{align}
Trained jointly with the HDR reconstruction network, this adapter effectively tailors the pretrained optical flow to HDR scenes and occlusion regions,  yielding flows with sharper and more reliable motion boundaries. Since the adapter is trained with normalized HDR inputs, it adapts to the HDR scenes well. After going through the flow adapter, we obtain the refined flow $\tilde{f}_{t\rightarrow t-1}$ and $\tilde{f}_{t\rightarrow t+1}$. 

After obtaining the refined flow maps, we warp the LDR inputs and fuse them together to generate a coarse HDR result. Specifically, we map the LDR inputs to HDR domain via Eq.~\ref{eq:ldr_to_hdr}. Then, $I_{t-1}$ and $I_{t+1}$ are aligned with $I_{t}$ using the estimated optical flows with the backward warping operation, resulting in $\tilde{I}_{t-1}$ and $\tilde{I}_{t+1}$. Considering the original $I_{t-1}$ and $I_{t+1}$ are beneficial for fusion in static regions, we also feed them into the fusion network. In summary, $\{\tilde{I}_{t-1}, \tilde{I}_{t+1}, I_t, I_{t+1},I_{t-1}\}$ are concatenated and fed into the fusion network. 

The fusion network follows a U-Net style encoder--decoder structure with skip connections. The encoder progressively extracts features at multiple scales using residual blocks, while the decoder reconstructs the spatial resolution through upsampling layers. The network outputs five weight maps $\{W_i\}_{i=1}^5$ corresponding to the five HDR frames used for fusion. The initial HDR reconstruction is obtained via weighted averaging:
\begin{align}
H_{\text{init}} = \frac{\sum_{i=1}^5 W_i \odot I_i}{\sum_{i=1}^5 W_i + \epsilon},
\end{align}
where $\odot$ denotes element-wise multiplication and $\epsilon=10^{-8}$ ensures numerical stability.

\begin{figure*}[t]
    \centering
    \includegraphics[width=0.9\linewidth]{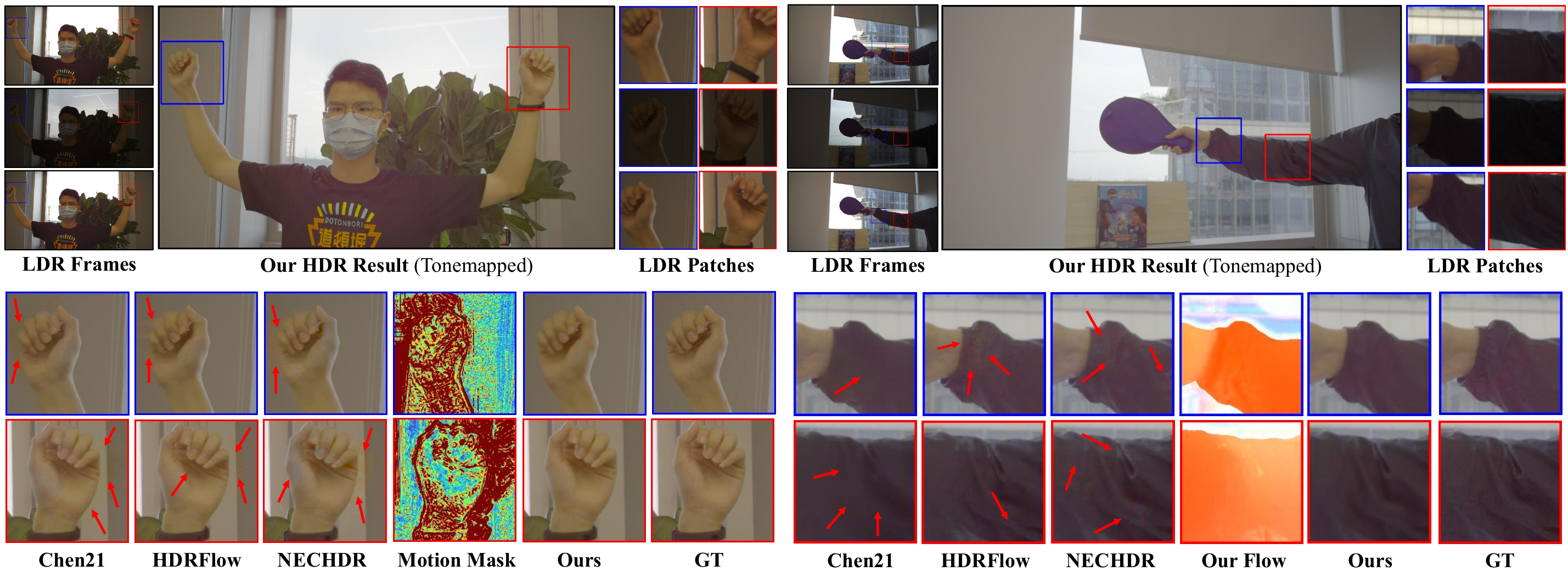}
    \vspace{-8pt}
    \caption{Qualitative comparisons with state-of-the-art methods on DeepHDRVideo dataset~\cite{chen2021hdr}. 
    }
    \vspace{-10pt}
    \label{fig:result_deaphdrvideo_compare}
\end{figure*}

\begin{figure*}[t]
    \centering
    \includegraphics[width=0.9\linewidth]{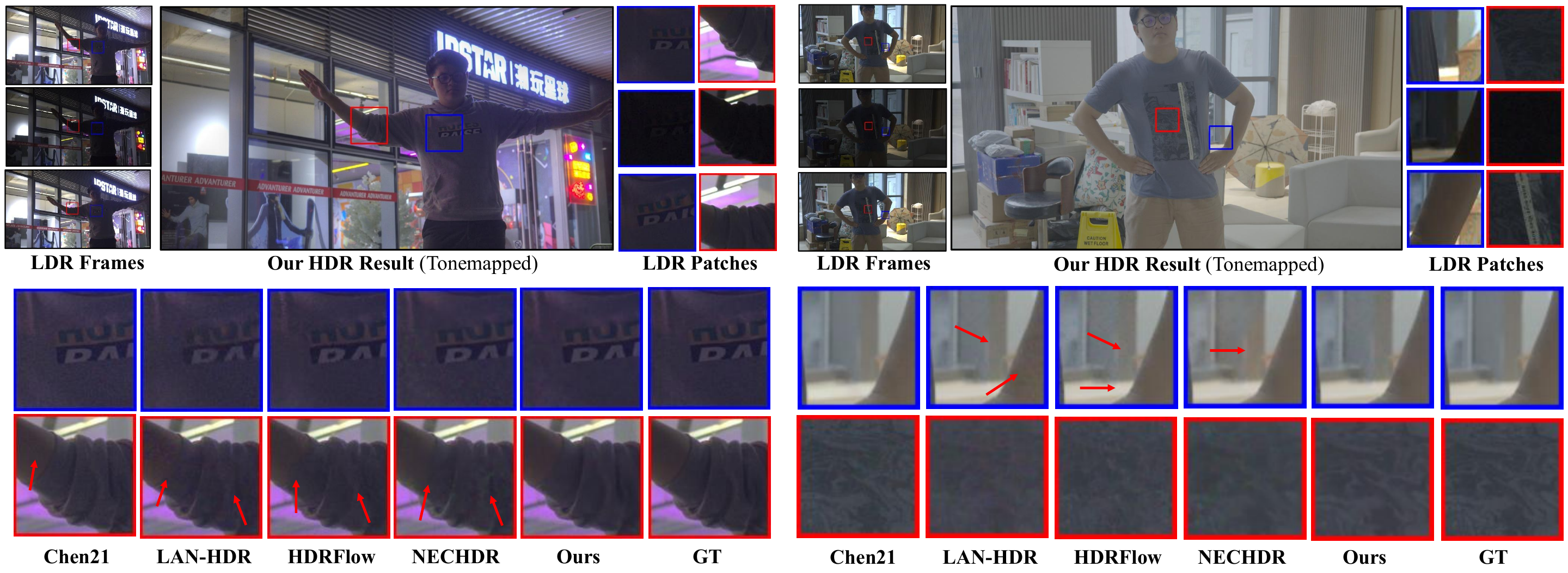}
    \vspace{-8pt}
    \caption{Qualitative comparisons with state-of-the-art methods on Real-HDRV dataset~\cite{shu2024towards}.
    }
    \vspace{-10pt}
    \label{fig:result_real_hdrv_compare}
\end{figure*}

\subsubsection{Physical Motion Modeling}
The ghosting and noise artifacts usually occur in the regions with motion. Therefore, it is important to  extract motion information from the input LDRs. Prior approaches include deriving occlusion masks from optical flow \cite{chen2025ultrafusion} or segmenting motion masks using a pretrained Segment Anything model \cite{yan2024dynamic}. However, these masks are binary and fail to capture the continuous physical characteristics of motion. Different from them, we propose to derive the motion mask from the physical decomposition of the optical flow field.
Given an optical flow $f=[u,v]$, we first compute its first-order spatial derivatives using Sobel operators:
\begin{align}
u_x &= \frac{\partial u}{\partial x}, \quad
u_y = \frac{\partial u}{\partial y}, \nonumber\\
v_x &= \frac{\partial v}{\partial x}, \quad
v_y = \frac{\partial v}{\partial y}.
\end{align}
These derivatives describe the local velocity gradient tensor $\nabla f$, from which we extract four physically interpretable motion components:
\begin{align}
\textbf{Translation: } &\;\; \|f\|_2 = \sqrt{u^2 + v^2}, \\
\textbf{Divergence: } &\;\; \nabla\!\cdot\!f = u_x + v_y, \\
\textbf{Curl: } &\;\; \nabla\!\times\!f = v_x - u_y, \\
\textbf{Shear: } &\;\; S = \tfrac{1}{2}(u_y + v_x), 
\end{align}
\iffalse
\begin{align}
\textbf{Translation: } &\;\; \|F\|_2 = \sqrt{u^2 + v^2}, 
&&\text{(global displacement magnitude)} \nonumber\\[2pt]
\textbf{Divergence: } &\;\; \nabla\!\cdot\!F = u_x + v_y, 
&&\text{(local expansion or compression)} \nonumber\\[2pt]
\textbf{Curl: } &\;\; \nabla\!\times\!F = v_x - u_y, 
&&\text{(local rotational motion)} \nonumber\\[2pt]
\textbf{Shear: } &\;\; S = \tfrac{1}{2}(u_y + v_x).
&&\text{(non-rigid deformation)}
\end{align}
\fi
These components correspond respectively to physical motion phenomena in the scene. Translation represents global displacement magnitude, divergence encodes scale changes, curl measures rotation, and shear captures local non-rigid deformation.
We then form a unified motion energy that integrates these physical cues:
\begin{equation}
\label{eq:E_m}
\scalebox{0.83}{$\displaystyle
E_m = 
\frac{
w_t \odot \|f\|_2 + 
w_d \odot |\nabla\!\cdot\!f| + 
w_c \odot |\nabla\!\times\!f| + 
w_s \odot |S|
}{
w_t + w_d + w_c + w_s + \epsilon
}.
$}
\end{equation}
Here $w_t, w_d, w_c, w_s$ are adaptive weights learned by a convolution block with the learned optical flow as input.  
This formulation ensures that the mask reflects not only the amplitude of motion but also the physical type and local structure of movement. For instance, a region with large curl but small divergence corresponds to pure rotation, while high divergence indicates zoom-like expansion. 

To further emphasize physically salient regions, a multi-scale contrast mechanism is applied:
\begin{align}
E_s = E_m \odot (1 + 2S_{\text{multi}}),
\end{align}
where $S_{\text{multi}}$ measures the center--surround motion contrast over scales $\{1,2,4\}$. In this way, the motion energy with large contrast (such as edges) can be enhanced. 
We then compute an adaptive Otsu-style threshold $\tau$ from the energy histogram to distinguish significant motion from background drift, and derive a soft motion mask:
\begin{align}
M = \frac{1}{2}\!\left[1 + \tanh\!\big(8(E_s - \tau)\big)\right].
\end{align}
The resulting $M$ is a continuous value field normalized to $[0,1]$, preserving spatial coherence while maintaining differentiability for end-to-end training.

This physically inspired construction ensures the mask captures genuine motion semantics, not just amplitude cues.
It highlights physically inconsistent regions (e.g., occlusion boundaries, reflections, moving objects) where ghosting arises, guiding second-stage fusion to adaptively suppress artifacts while preserving spatial consistency.

\subsection{Stage II: Refinement Stage}
\subsubsection{Motion Guided Feature Modulation}
While the fusion network produces a reasonable initial HDR result, residual ghosting artifacts and noise may persist in regions with complex motion or occlusions. To address this, we introduce the refinement stage. 

Considering that weighted fusion of the original input is good at preserving the image structures but cannot eliminate artifacts, such as ghosting and noise, in this stage, different from the coarse fusion stage, we propose to process the inputs in feature domain. Specifically, for each input frame $L_t$ and its linear HDR counterpart $I_t$, we extract its features via an encoder $\mathcal{E}$, generating $F_{L_t}$ and $F_{I_t}$. The two features are then fused together via a $1\times1$ convolution, namely 
\begin{align}
F_{LI_t} = \text{Conv}_{1\times1}\left([F_{L_t}, F_{I_t}]\right). 
\end{align}
Then, the neighboring frame features $F_{LI_{t-1}}$ and  $F_{LI_{t+1}}$ are aligned with the central frame feature using the optical flow $\tilde{f}_{t\rightarrow t-1}$ and $\tilde{f}_{t\rightarrow t+1}$, generating $\tilde{F}_{LI_{t-1}}$ and $\tilde{F}_{LI_{t+1}}$. A straightforward solution is directly fusing the aligned feature with $F_{LI_t}$. However, this cannot make the network focus on the regions with artifacts. Therefore, we further propose to encode the motion information and use the motion features to modulate the aligned features. Specifically,  the motion feature is obtained by 
\begin{align}
G_{t\pm 1} = \sigma\left(\text{Conv}\left([\|\tilde{f}_{t\rightarrow t\pm 1}\|_2/\tilde{f}_\text{max}, M_{t\pm 1}, \mathbf{1}]\right)\right),
\end{align}
where Conv is a convolution block, $\tilde{f}_\text{max}$ is the maximum flow magnitude, and 1 is a constant channel that provides bias information. 
$\sigma$ is the sigmoid function, which maps the motion features into the range [0,1]. Then $G_{t\pm 1}$ is used to modulate the aligned features via 
\begin{align}
\bar{F}_{LI_{t\pm1}} = G_{t\pm 1} \odot \tilde{F}_{LI_{t\pm1}}
\end{align}
Therefore, the regions with obvious motions will be emphasized and the static regions will be suppressed. 

\subsubsection{Feature Enhancement and Fusion}
After obtaining the modulated features, we further combine it with the central frame feature $F_{LI_t}$. Then, these features go through a feature enhancement module, which is constructed by three dilated branches.  
In this way, we can use correlations in a large receptive field to reduce noise and artifacts. Then we fuse the features extracted from the coarse fusion result $H_\text{coarse}$ obtained in the first stage with these enhanced features. Finally, the fused features go through the decoder, reconstructing the final HDR result $H_\text{final}$.

\subsubsection{Training Loss}
Following previous works~\cite{kalantari19,yan2019attention,wu2018deep}, we use the differentiable $\mu$-law function as the tonemapping function $\mathcal{T}$:
\begin{equation} 
\mathcal{T}(H) = \frac{\log(1+\mu H)}{\log(1+\mu)},
\end{equation}
where $\mu$ is set to 5,000. 
We compute the reconstruction loss $\mathcal{L}$ between the predicted ${H}_{\text{final}}$ and ground-truth ${H}_{\text{gt}}$ using $\mathcal{L}_1$ loss:
\begin{equation} 
\begin{aligned}
\mathcal{L} =  \parallel \mathcal{T}({H}_{\text{final}}) - \mathcal{T}({H}_{\text{gt}}) \parallel_1.
\end{aligned}
\end{equation}
\section{Experiments}
\label{sec:experiment}

\subsection{Experimental Setup}
\textbf{Datasets.} 
To ensure a fair comparison, we also adopt the Vimeo-90K~\cite{xue2019video} dataset as training data. Following prior works~\cite{xu2024hdrflow,chen2021hdr,cui2024exposure,lan_hdr}, we generate the corresponding alternating-exposure LDR sequences through a simulation pipeline. During training, we apply random flipping and rotation, followed by randomly cropping the augmented frames to 256$\times$256 patches as network inputs.

\begin{table}[t]
\setlength{\tabcolsep}{3pt}
\centering
\caption{Quantitative comparisons on the DeepHDRVideo dataset~\cite{chen2021hdr}, including both image-based~\cite{kalantari13, yan2019attention,prabhakar,liu2022ghost} and video-based methods~\cite{kalantari19,chen2021hdr,lan_hdr,cui2024exposure,xu2024hdrflow,guan2024diffusion}. * indicates additional training with the Sintel dataset. \best{Red Bold}: best, \second{Blue Bold}: second best.}
\vspace{-8pt}
\resizebox{1.0\linewidth}{!}{%
\begin{tabular}{
    l
    >{\centering\arraybackslash\scriptsize}p{1.0cm}
    >{\centering\arraybackslash}m{1.2cm}
    >{\centering\arraybackslash}m{1.2cm}
    >{\centering\arraybackslash}m{1.1cm}
}
\specialrule{0.12em}{0pt}{1pt}
\textbf{Methods} &  & \textbf{PSNR\textsubscript{$T$}} & \textbf{SSIM\textsubscript{$T$}} & \textbf{HDR-VDP-2} \\
\specialrule{0.06em}{1pt}{1pt}

% --------- Four-row merged cell starts here ---------
Kalantari13~\cite{kalantari13}      & TOG2013   & 40.33 & 0.9409 & 
\multirow{4}{*}{\centering --} \\

AHDRNet~\cite{yan2019attention}     & CVPR2019  & 40.54 & 0.9452 & \\

Prabhakar~\cite{prabhakar}          & CVPR2021  & 40.21 & 0.9414 & \\

HDR-Transformer~\cite{liu2022ghost} & ECCV2022  & 41.51 & 0.9458 & \\
% --------- Four-row merged cell ends here ---------

\specialrule{0.03em}{1pt}{1pt}

Kalantari19~\cite{kalantari19}      & CGF2019   & 39.91 & 0.9329 & -- \\
Chen21~\cite{chen2021hdr}           & ICCV2021  & 43.32 & 0.9551         & \second{78.37} \\
LAN-HDR~\cite{lan_hdr}              & ICCV2023  & 41.83          & 0.9499         & 76.00 \\
NECHDR~\cite{cui2024exposure} & MM24 & \second{43.44}          & 0.9558 & 77.28 \\
HDRFlow~\cite{xu2024hdrflow}        & CVPR2024  & 43.03          & 0.9518         & 77.58 \\
HDRFlow*~\cite{xu2024hdrflow} & CVPR2024 & 43.25  & 0.9515         & 78.07 \\
HDR-V-Diff~\cite{guan2024diffusion} & ECCV2024  & 42.07          & \best{0.9604}  & -- \\
\specialrule{0.06em}{1pt}{1pt}
\textbf{Ours}                       &           & \best{43.87}   & \second{0.9573} & \best{78.88} \\
\specialrule{0.12em}{1pt}{0pt}
\end{tabular}
}
\vspace{-8pt}
\label{tab:Chen_Dataset_Full_List}
\end{table}

\begin{table}[t]
\setlength{\tabcolsep}{1.5pt} 
\centering
\caption{Quantitative comparison of different methods on the Real-HDRV dataset~\cite{shu2024towards}. * indicates additional training with the Sintel dataset. \best{Red Bold}: best, \second{Blue Bold}: second best.}
\vspace{-8pt}
\resizebox{1.0\linewidth}{!}{
\begin{tabular}{lcccccc}
\specialrule{0.12em}{0pt}{2pt}
Methods 
& \textbf{PSNR\textsubscript{$T$}} 
& \textbf{SSIM\textsubscript{$T$}} 
& \textbf{HDR-VDP-2}
& PSNR\textsubscript{$L$} 
& SSIM\textsubscript{$L$} \\
\specialrule{0.06em}{2pt}{2pt}
Chen21~\cite{chen2021hdr}                        & 40.79 & 0.9510 & 76.50 & 50.30 & 0.9943 \\
LAN-HDR~\cite{lan_hdr}                           & 39.79 & 0.9448 & 74.43 & 46.54 & 0.9928 \\
HDRFlow~\cite{xu2024hdrflow}                     & 40.34 & 0.9481 & 75.82 & 49.72 & 0.9941 \\
\makecell{HDRFlow*}~\cite{xu2024hdrflow} & 40.28 & 0.9450 & 76.08 & 49.87 & 0.9941 \\
NECHDR~\cite{cui2024exposure}                    & \second{40.88} & \second{0.9518} & \second{76.77} &  \second{50.41}& \best{0.9952} \\
\specialrule{0.06em}{2pt}{2pt}
\textbf{Ours}                                    & \best{41.01} & \best{0.9538} & \best{76.93} & \best{50.51} & \second{0.9949} \\
\specialrule{0.12em}{2pt}{0pt}
\end{tabular}}
\label{tab:realhdrv_result}
\vspace{-10pt}
\end{table}

To evaluate the models’ capability of reconstructing HDR content from real-world data, we adopt the two largest real-captured HDR video benchmarks as our test set, namely DeepHDRVideo~\cite{chen2021hdr} and Real-HDRV~\cite{shu2024towards}. DeepHDRVideo contains 76 dynamic scenes and 49 static scenes. For the Real-HDRV dataset, we utilize the 50 test videos released by \cite{shu2024towards}. Since the official testing code and evaluation-oriented data organization of Real-HDRV are not released, we follow a unified protocol by using a sliding-window strategy to traverse and evaluate all frames within each selected video. For each frame, we utilize its two neighbors to assist the HDR reconstruction.

\noindent\textbf{Implementation details.} We implement our method in PyTorch and conduct all experiments on an NVIDIA RTX 3090 GPU. We employ the Adam optimizer~\cite{adam} with $\beta_1=0.9$ and $\beta_2=0.999$. The initial learning rate is set to $1\times10^{-4}$ and is reduced by half every 10 epochs. We set the $\mu$ parameter of the $\mu$-law tone mapping function to 5000. We train our model with 50 epochs using a batch size of 16. 

\noindent\textbf{Evaluation metrics.}  
We adopt PSNR\textsubscript{$T$}, SSIM\textsubscript{$T$}, 
PSNR\textsubscript{$L$}, SSIM\textsubscript{$L$}, and HDR-VDP-2 as our evaluation metrics. Among them, PSNR\textsubscript{$T$} and SSIM\textsubscript{$T$} are computed on the $\mu$-law tonemapped HDR domain, 
while PSNR\textsubscript{$L$} and SSIM\textsubscript{$L$} are computed 
in the linear HDR domain. For a fair comparison, all HDR-VDP-2 results in this paper 
are obtained using a fixed evaluation configuration, where the display diagonal is 
set to 28 inches and the viewing distance to 55 cm.

\subsection{Comparisons with State-of-the-art}

\begin{table}[t]
\centering
\caption{Comparison of optical flow accuracy on Real-HDRV~\cite{shu2024towards}.}
\vspace{-8pt}
\resizebox{1.0\columnwidth}{!}{
\setlength{\tabcolsep}{8pt}
\renewcommand{\arraystretch}{1.0}
\small
\begin{tabular}{l c c c c}
\toprule
& \textbf{HDRFlow} & \textbf{NECHDR} & \textbf{Ours w/o FA} & \textbf{Ours} \\
\midrule
Optical Flow EPE $\downarrow$   & 3.53 & 3.30 & \best{1.81} & \second{2.08} \\
LDR Warping PSNR $\uparrow$  &   29.93   &   30.31   &  \second{30.39}  & \best{30.84}  \\
LDR Warping SSIM $\uparrow$  &    0.8763   & \second{0.8901}    &  0.8897 &  \best{0.8922}  \\
\bottomrule
\end{tabular}
}

\label{tab:flow_align_compare}
\end{table}

\begin{figure}
\centering
\vspace{-10pt}
{\includegraphics[width=1.0\linewidth]{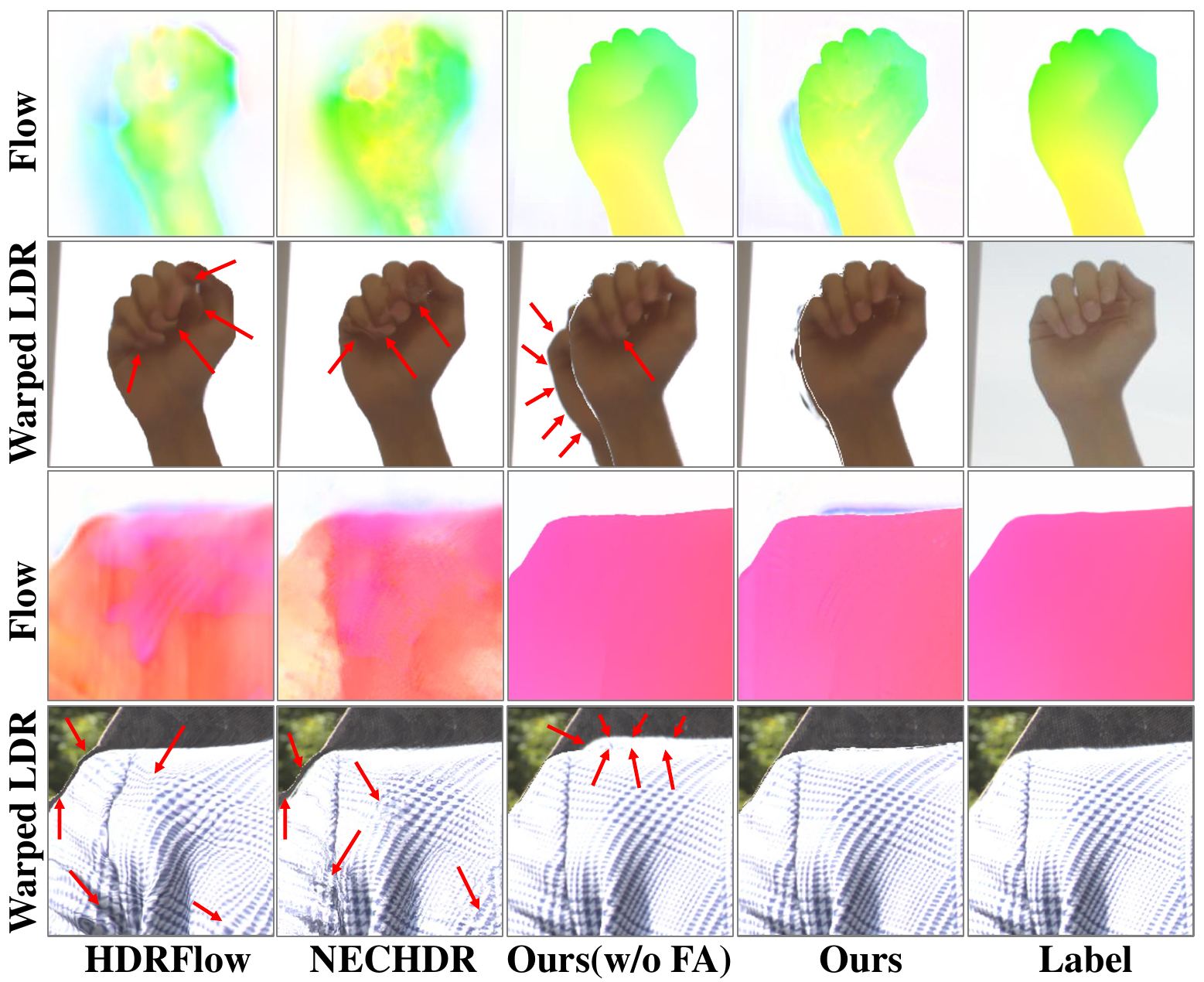}}
\caption{Qualitative Comparisons of optical flow and warped LDR on DeepHDRVideo (Rows 1-2)~\cite{chen2021hdr} and Real-HDRV (Rows 3-4)~\cite{shu2024towards} Datasets.}
\vspace{-15pt}
\label{fig:warp_comparison}
\end{figure}

\noindent\textbf{Quantitative evaluations.} 
We compare our approach with state-of-the-art (SOTA) methods. Since many image HDR reconstruction methods also evaluate on the DeepHDRVideo dataset, we directly quote their results from their papers \cite{kalantari13, yan2019attention,prabhakar,liu2022ghost}. For the video HDR reconstruction methods, since there are no released models for \cite{kalantari19}, we quote its results from \cite{xu2024hdrflow}. Since HDR-V-Diff~\cite{guan2024diffusion} does not release its code, we are unable to evaluate its HDR-VDP-2~\cite{hdr_vap_2} performance under our experimental settings. We directly quote its PSNR and SSIM results from \cite{guan2024diffusion}.  

As shown in Tab.~\ref{tab:Chen_Dataset_Full_List}, our model achieves the best PSNR\textsubscript{$T$} and best HDR-VDP-2 scores on the DeepHDRVideo dataset, surpassing the second-best method by 0.43 dB in PSNR\textsubscript{$T$}. On the Real-HDRV dataset, as shown in Tab.~\ref{tab:realhdrv_result}, our method obtains the highest PSNR\textsubscript{$T$}, SSIM\textsubscript{$T$}, HDR-VDP-2, and PSNR\textsubscript{$L$} scores among all approaches.

\noindent\textbf{Qualitative evaluations.} 
We compare the visual results of our approach with state-of-the-art methods~\cite{chen2021hdr,xu2024hdrflow,cui2024exposure,lan_hdr} in Fig. \ref{fig:result_deaphdrvideo_compare} and Fig. \ref{fig:result_real_hdrv_compare}.
As shown in Fig.~\ref{fig:result_deaphdrvideo_compare}, our method produces significantly better visual quality. Our motion mask effectively highlights motion-induced regions and suppresses ghosting artifacts, whereas all other methods exhibit varying degrees of ghosting in these challenging areas.
As shown in Fig.~\ref{fig:result_real_hdrv_compare}, our approach delivers the most robust performance in challenging night time scenes. Our reconstruction contains the least noise and preserves the most faithful texture details, while compared methods suffer from noise amplification inherited from the original LDR frames and exhibit noticeable texture degradation.

Tab. \ref{Tab:time} presents the inference time of different methods. Our method is much faster than \cite{chen2021hdr} and \cite{lan_hdr}.  

\noindent\textbf{Evaluation of flow.} 
Since there is no GT flow for HDR task, we generate pseudo flow labels by estimating flow with HDR ground truth frames as input. We adopt MemFlow~\cite{dong2024memflow} for pseudo label generation, as using an independent third-party optical flow network avoids evaluation bias and maximizes assessment reliability. For evaluation on the Real-HDRV Dataset~\cite{shu2024towards}, we first use EPE (end-point error) between estimated flow and label. Considering that warping quality is essential for the fusion process, we further evaluate the estimated flow quality between the warped and target LDR frames at the same exposure. Our flow estimation method achieves lower EPE and the best warping quality in Tab. \ref{tab:flow_align_compare}. Note that, after introducing flow adapter (FA), the EPE is increased but warping quality is improved. The reason is that our FA can generate offsets for occlusion regions, which is beneficial for warping. However, the pseudo flow label for these regions is zero, see Fig.~\ref{fig:warp_comparison}. Note that , since there is only one LDR exposure for each frame in the DeepHDRVideo dataset~\cite{chen2021hdr}, we present the HDR ground truth as the label (Fig.~\ref{fig:warp_comparison}), which is only for qualitative evaluation.

\begin{table}[t]
\centering
\caption{Inference time comparison for two different resolutions evaluated on an NVIDIA RTX 3090 GPU.}
\vspace{-8pt}
\renewcommand{\arraystretch}{1.2}
\setlength{\tabcolsep}{3pt} 
\scalebox{0.8}{
\begin{tabular}{ccc}
\specialrule{0.12em}{0pt}{2pt}
\multirow{2}{*}{Methods} 
& \multicolumn{2}{c}{\textbf{Time (s)}} \\
\cmidrule(lr){2-3}
& \textbf{1920×1080} & \textbf{1536×813} \\
\specialrule{0.06em}{2pt}{2pt}
Kalantari19~\cite{kalantari19}  & 0.260 & 0.220 \\
HDRFlow~\cite{xu2024hdrflow}    & 0.076 & 0.050 \\
NECHDR~\cite{cui2024exposure}   & 0.160 & 0.103 \\
Chen21~\cite{chen2021hdr}       & 0.570 & 0.540 \\
LAN-HDR~\cite{lan_hdr}          & 0.905 & 0.525 \\
Ours                             & 0.290 & 0.190 \\
\specialrule{0.12em}{2pt}{0pt}
\end{tabular}
}
\label{Tab:time}
\vspace{-10pt}
\end{table}

\subsection{Ablation Study}
\begin{figure}[t]
    \centering
    \includegraphics[width=1.0\linewidth]{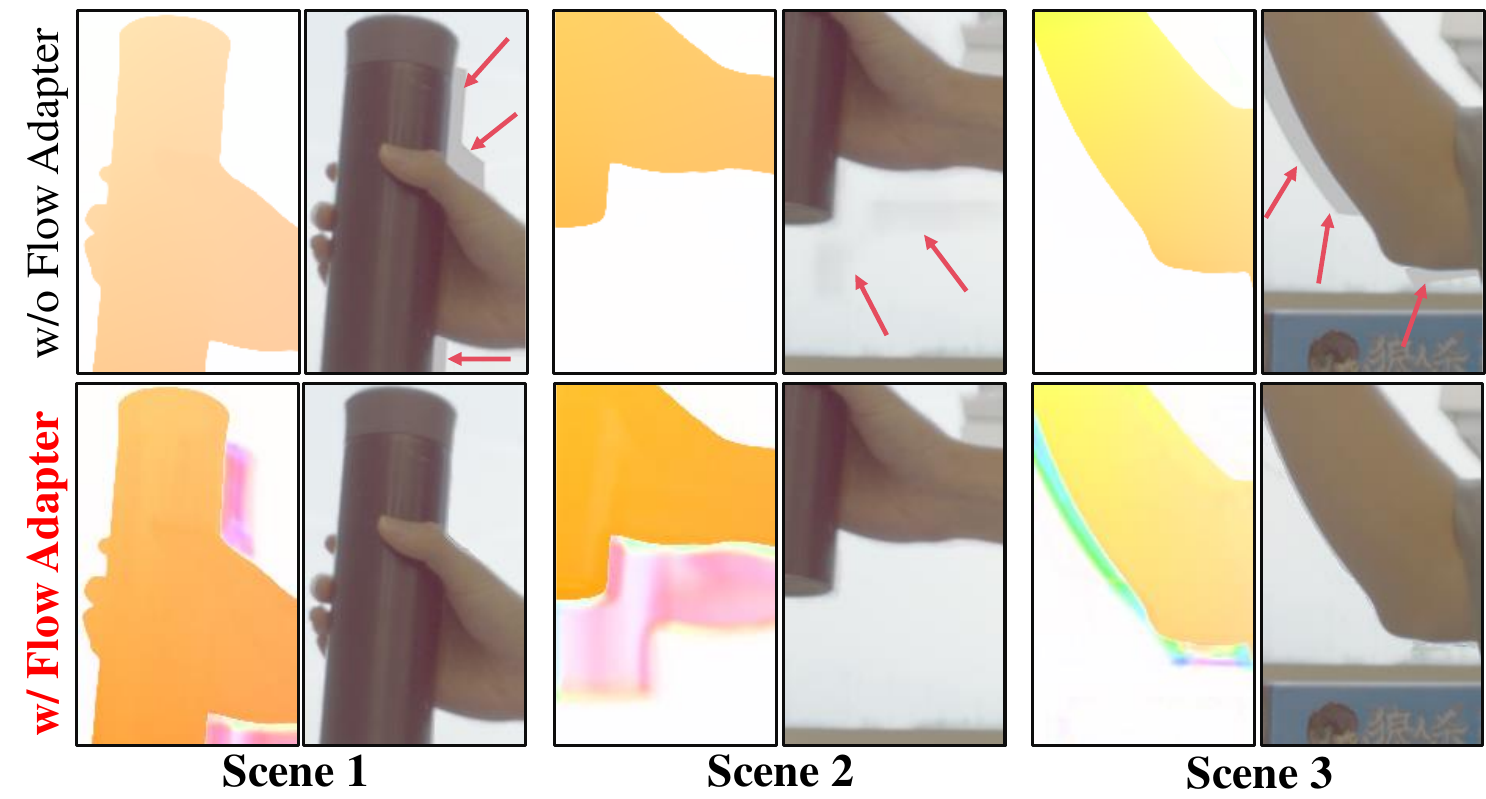}
    \vspace{-15pt}
    \caption{Ablation of flow adapter by removing it from our full solution. 
    }
    \label{fig:ablation}
    \vspace{-15pt}
\end{figure}

We conduct ablation studies on the DeepHDRVideo dataset and the Real-HDRV dataset. The quantitative results are summarized in Tab.~\ref{tab:ablation study}. We first present the first stage result without flow adapter. The result is the worst, since the flow adapter is essential, as it make the pretrained flow models adapt to HDR scenarios. Then we introduce the second stage to the first variant. The result is improved, but is still much worse than our full solution. The main reason is that a high quality flow map is  a solid foundation for the refinement stage. As shown in Fig.~\ref{fig:ablation}, the visual results further verify the importance of the Flow Adapter in reducing ghosting.

Then, we evaluate the effect of our motion mask. The result is degraded by 0.19 dB on the DeepHDRVideo dataset if we remove the motion mask from our full solution. The main reason is that the  motion mask helps the refinement network aggregate the most informative features, leading to more accurate HDR reconstruction. Replacing our physics motion modeling with a standard learned CNN also leads to inferior results. In addition, compared with single stage solution, our two stage solution greatly improves the reconstruction performance. Please refer to our supplementary file for more details.

\begingroup
\renewcommand{\arraystretch}{1.2}
\begin{table}[t]
\caption{Ablation study of $\text{F}^2\text{HDR}$ on DeepHDRVideo dataset and Real-HDRV dataset. FA denotes Flow Adapter. Mask* denotes the mask learned via a CNN.}
\vspace{-8pt}
\scalebox{0.9}{
\setlength{\tabcolsep}{1.2pt} 
\centering
\begin{tabular}{lcccc}
\specialrule{0.12em}{0pt}{2pt}
\multirow{2}{*}{Methods} 
& \multicolumn{2}{c}{\textbf{DeepHDRVideo}} 
& \multicolumn{2}{c}{\textbf{Real-HDRV}} \\
\cmidrule(lr){2-3} \cmidrule(lr){4-5}
& PSNR\textsubscript{$T$} & SSIM\textsubscript{$T$} & PSNR\textsubscript{$T$} & SSIM\textsubscript{$T$} \\
\specialrule{0.06em}{2pt}{2pt}
Stage I w/o FA              & 43.06 & 0.9514 & 40.29 & 0.9453 \\
Stage I w/o FA + Stage II       & 43.21 & 0.9530 & 40.45 & 0.9461 \\
Stage I               & 43.39 & 0.9549 & 40.61 & 0.9481 \\
Stage I w/o Mask + Stage II       & 43.68 & 0.9565 & 40.87 & 0.9502 \\
Stage I w/  Mask* + Stage II       & 43.76 & 0.9569 & 40.93 & 0.9524 \\
Ours (Stage I + Stage II)   & \textbf{43.87} & \textbf{0.9573} & \textbf{41.01} & \textbf{0.9538} \\
\specialrule{0.12em}{2pt}{0pt}
\end{tabular}
}
\label{tab:ablation study}
\end{table}
\vspace{-8pt}
\endgroup

\section{Conclusion}
In this paper, we presented \textit{F\textsuperscript{2}HDR}, a two-stage HDR video reconstruction framework that addresses the fundamental challenges of cross-exposure misalignment and motion-induced artifacts in dynamic scenes. Our design combines two key components: a \emph{Flow Adapter} that transforms generic pretrained optical flow into task-adaptive cross-exposure motion fields, and a \emph{Physical Motion Modeling} module that extracts continuous, physically meaningful motion cues to guide the refinement process. By integrating these components within a coarse-to-fine reconstruction pipeline, our method effectively aggregates complementary information from both LDR and HDR domains. Experiments demonstrate the superiority of our method.

{
    \small
    \bibliographystyle{ieeenat_fullname}
    \bibliography{main}
}
\end{document}